\documentclass{article}
\usepackage{color}
\usepackage{graphicx}
\usepackage{subfigure}
\usepackage{caption}
\usepackage{times}
\usepackage{amsfonts}
\usepackage[switch]{lineno}
\usepackage{amsmath}
\usepackage{multirow}
\usepackage{cprotect}
\usepackage[colorlinks=true,pagebackref=false,citecolor=black]{hyperref}
\usepackage{booktabs}
\usepackage{ijcai}
\usepackage{amssymb}

\usepackage{etoolbox}
\makeatletter
\patchcmd{\maketitle}
{\def\@makefnmark}
{\def\@makefnmark{}\def\useless@macro}
{}{}
\makeatother

\title{Learning Deep Features via Congenerous Cosine Loss for  Person Recognition
	}
\author{Yu Liu\textnormal{\textsuperscript{1*},} Hongyang Li\textnormal{\textsuperscript{2*} and }Xiaogang Wang\textnormal{\textsuperscript{2}}\\ 
\textsuperscript{1} SenseTime Group Ltd., Beijing, China  \\
\textsuperscript{2} The Chinese University of Hong Kong, New Territories, Hong Kong  \\
liuyu@sensetime.com, \{yangli, xgwang\}@ee.cuhk.edu.hk
\thanks{* The first two authors contribute equally.}
}
\begin{document}
\pagenumbering{roman}
\maketitle

\begin{abstract}
 Person recognition  aims at recognizing the same identity across time and space with complicated scenes and similar appearance. 
 In this paper, we propose a novel method to address this  
 task by training a network to obtain robust and representative features.
 %
\textcolor{black}{The intuition is that we directly compare and optimize the cosine distance between two features - enlarging inter-class distinction as well as alleviating inner-class variance.
 We propose a congenerous cosine loss
 by minimizing the cosine distance between samples and their cluster centroid in a cooperative way. 
 Such a design reduces the complexity and could be implemented via softmax with normalized inputs.} Our method also differs from previous work in person recognition 
 that
 we do not conduct a second training on the test subset.
 %
 %
 The identity of a person is determined by measuring the similarity from several body regions in the reference set. 
 %
 %
 %
 Experimental results show that the proposed approach achieves better  classification accuracy against previous state-of-the-arts. 
\end{abstract}

\section{Introduction}\label{sec:intro}

With an increasing 
demand of intelligent cellphones and digital cameras, people today take more photos to jot down 
daily life and stories. Such an overwhelming trend is generating a desperate demand for smart tools to recognize the same person (known as \textit{query}), across different time and space, among thousands of images from personal data, social media or Internet.
\begin{figure}[t]
	\centering
	\includegraphics[width=.43\textwidth]{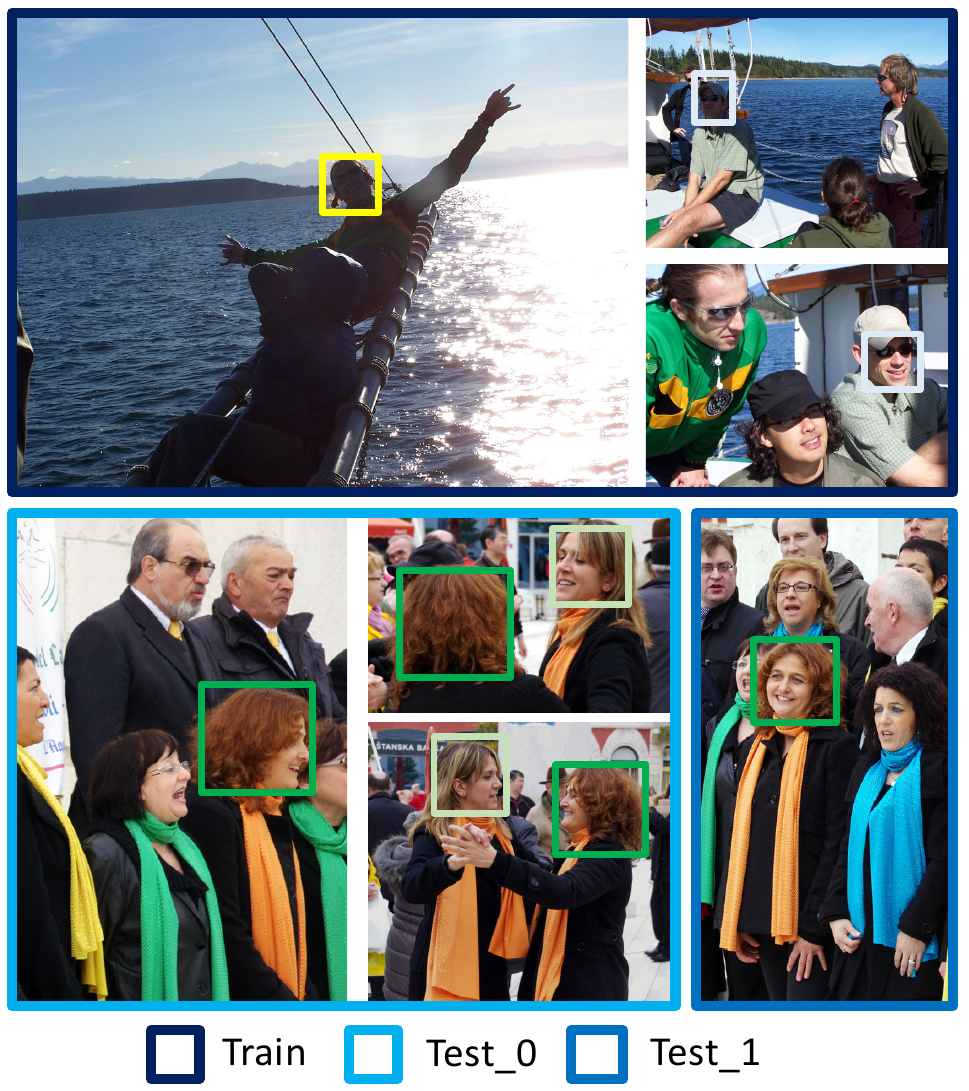}
	\caption{Overview of the PIPA dataset. 
		Previous work use the 
		\texttt{train} 
		set only for finetuning features; the recognition system is trained on \texttt{test\_0}
		and evaluated on 
		\texttt{test\_1}. In our work, we directly train the model on \texttt{train} and use \texttt{test\_0} as reference to predict the identities on \texttt{test\_1} by measuring the feature similarity between two sets.
	}
	\label{fig:dataset}
\end{figure}
Previous work \cite{original,oh_eccv,oh_iccv,person_recog_rnn,piper,zlin} has demonstrated that  person recognition in such unconstrained settings remains a challenging problem due to many factors, such as non-frontal faces, varying light and illumination, the variability in appearance, texture of identities, 
etc.

The recently proposed PIPA dataset \cite{piper} contains thousands of images with complicated scenarios and similar appearance among persons. The illumination, scale and context of the data varies a lot and many instances have partial or even no faces.
Figure \ref{fig:dataset} shows some samples from both the training and test sets.
Previous work \cite{piper,oh_iccv,zlin} resort to identifying the same person via a multi-cue, multi-level manner where the training set is used only for extracting features and the follow-up classifier (SVM or neural network) is trained on the \texttt{test\_0} set\footnote{ We denote \texttt{test\_0} as the reference set throughout the paper.}. The recognition system is evaluated on the \texttt{test\_1} set. We argue that 
such a practice is infeasible and ad hoc in realistic application since the second training on  \texttt{test\_0}   is auxiliary and needs re-training if new samples are added. Instead, we aim at providing a set of robust and well generalized feature representations, which is trained directly on the training set, and at identifying the person by measuring the feature similarity between two splits on test set. There is no need to train on 
\texttt{test\_0} and the system still performs well even if new data comes in.

As shown in the bottom of Figure \ref{fig:dataset}, the woman  in green box wears similar  scarf as does the person in light green box. Her face is shown partially or does not appear in some cases. To obtain robust feature representations, we train several deep models for different regions and combine the similarity score of features from different regions to have the  prediction of one's identity.
Our key observation is that during training, the cross-entropy loss does not 
guarantee the similarity among samples \textit{within} a category. It magnifies the difference across classes and ignore the feature similarity of the same class.
To this end, we 
propose a \textit{congenerous cosine loss}\footnote{  \href{https://github.com/sciencefans/coco_loss}{\texttt{https://github.com/sciencefans/coco\_loss}}}, namely \textbf{COCO}, to enlarge the inter-class distinction as well as narrow down the inner-class variation. It is achieved by measuring the cosine distance between sample and its cluster centroid in a cooperative manner.
Moreover, we also align each region patch to a pre-defined base location to further make samples within a category
be more closer in the feature space. Such an alignment strategy could also 
make the network less prone to overfitting.

\begin{figure*}[t]
	\centering
	\includegraphics[width=.98\textwidth]{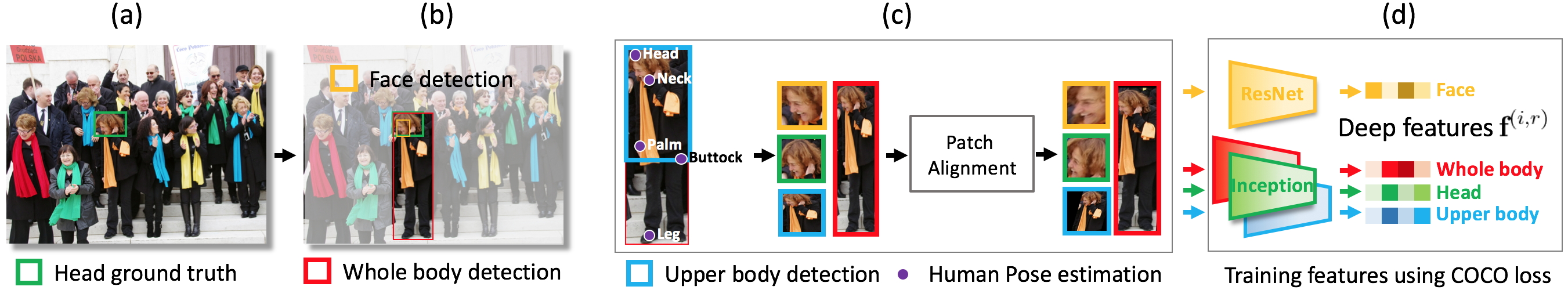}
	\vspace{-.1cm}
	\cprotect\caption{Training workflow of our algorithm. 
		(a) Each person is labelled with a ground-truth box of the head. 
		(b)-(c): Face, whole body and upper body detection.
		Then, we align each region (patch) to a base position to alleviate the inner-class variance. (d) Each aligned patch is further fed into a deep model to obtain representative and robust features using COCO loss.
	}
	\label{fig:pipeline}
\end{figure*}

Figure \ref{fig:pipeline} illustrates the training pipeline of our proposed algorithm at a glance. Each instance in the image is annotated with a ground truth head and we train a face and human body detector respectively, using the RPN framework \cite{faster_rcnn} to detect these two regions. Then a human pose estimator \cite{pose} is applied to detect key parts of the person in order to localize the upper body region. After cropping four region patches, we conduct an affine transformation to align different patches from training samples to a `base' location.
Four deep models are trained separately on the PIPA training set using the COCO loss to obtain a set of robust features. To sum up, the contributions in this work are as follows:
\begin{itemize}
	\item 	\textcolor{black}{Propose a congenerous cosine loss to directly optimize the cosine distance among samples within and across categories. It is achieved in a cheap softmax manner with normalized inputs and less complexity.}
	\item \textcolor{black}{Design a person recognition pipeline that leverages from  several body regions to obtain a discriminative representation of  features, without the necessity of conducting a second training on the test set.}
	\item Align region patches to the base location via affine transformation to reduce variation among samples, making the network less prone to overfitting.
\end{itemize}

\section{Related Work}
\textbf{Person recognition} in photo albums \cite{original,oh_eccv,oh_iccv,person_recog_rnn,piper,zlin} aims at recognizing the identity of people in daily life photos, where such scenarios can be complex with cluttered background.
%
\cite{original} first address the problem by proposing a Markov random filed framework to combine all contextual cues to recognize the identity of persons.
Recently, 
\cite{piper} introduce a large-scale dataset for this task.
They accumulate the cues of poselet-level person recognizer trained by a deep model to compensate for pose variations.
%
%
%
%
In \cite{oh_iccv}, a detailed analysis of different cues 
is explicitly investigated and three additional test  splits are proposed for evaluation. 
\cite{person_recog_rnn} embed scene and relation contexts in LSTM and formulate person recognition as a sequence prediction task.

\textbf{Person re-identification} is to 
match pedestrian images from various perspectives in
cameras for a typical time period and has led to many important applications in video \cite{reid_1,reid_2,reid_3,reid_4,reid_5}. Existing
work employ metric learning and
mid-level feature learning to address this problem. 
\cite{reid_1} 
propose a deep network using pairs of people to encode
photometric transformation. 
\cite{reid_2} incorporate a Siamese
deep network to learn the similarity metric between pairs of
images. The main difference between person recognition and re-identification resides in
the data logistics. The former is to identify the same person across places and time. 
In most cases, the identity varies a lot in appearance under different occasions. The latter is 
to detect person in a consecutive video, meaning that the appearance and background do not vary much in terms of time.

\textbf{Deep neural networks} \cite{alexnet,resNet} have dramatically advanced the computer vision community in recent years, 
with high performance boost in tremendous tasks, such as image classification
 \cite{
	resNet,li2016multi}, object detection \cite{fast_rcnn,li_zoom_network}, object tracking \cite{czz_tracking}, etc.
The essence behind the success of deep learning 
resides in both its superior expression power of non-linear complexity in high-dimension space \cite{hinton_nature}
and large-scale datasets \cite{imagenet_conf,celebrity} where the deep networks could, in full extent, learn complicated patterns and representative features.

\section{Algorithm}
\textcolor{black}{The proposed COCO loss is trained for each body region to obtain robust features. Section \ref{sec:face-detection-and-alignment} depicts the detection of each region given an image; for inference in Section \ref{sec:inference}, we extract the features from corresponding regions on both \texttt{test\_1} and \texttt{test\_0}, merge the similarity scores from different regions and make the final prediction on one's identity.}
\subsection{Region detection}\label{sec:face-detection-and-alignment}

The features of four regions $r\in\{1,\cdots,4\}$, namely, \textit{face}, \textit{head}, \textit{whole body} and \textit{upper body}, are utilized to train the features. We first state the detection of these regions.

\textbf{Face}. We  pre-train a face detector in a region proposal network (RPN) spirit following Faster RCNN \cite{faster_rcnn}. The source of  data comes from  Internet and the number of images is roughly 300,000. The network structure is a shallow version of the ResNet model \cite{resNet} where we remove layers after \texttt{res\_3b} and add two loss heads (classification and regression).
 Then we finetune the face model on PIPA  training set for COCO loss. The face detector identifies $m$ keypoints of the face (eye, brow, mouth, etc.) and we align the detected face patch to a `base' shape via translation, rotation and scaling. 
Let  $p, q \in \mathbb{R}^{m \times 2} $ denote  $m$ keypoints detected by the face model and the aligned results, respectively. We define $P,Q$ as two affine spaces, then an affine transformation
$\mathcal{A}: P \mapsto Q$ is defined as: 
\begin{equation}
p \mapsto q = \textbf{A} p +b, 
\end{equation}
where $\textbf{A }\in  \mathbb{R}^{m \times m} $ is a linear transformation matrix in $P$ and $b  \in  \mathbb{R}^{m \times 2} $ being the bias in $Q$.
%
\textcolor{black}{
Such an alignment scheme ensures samples both within and across category do not have large variance: if the network is learned without alignment, it has to distinguish more patterns, \textit{e.g.}, different rotations among persons, making it more prone to overfitting; if the network is equipped with alignment, it can focus more on differentiating features of different identities despite of rotation, viewpoint, translation, etc.
}

\textbf{Head, whole body and upper body.}
The head region is given as the ground truth for each person.
 To detect a whole body, we also pre-train a detector in the RPN framework.
The model is trained on the large-scale human celebrity dataset \cite{celebrity},
where we use the
first 87021 identities in 4638717  images.
The network structure is an inception model \cite{bn} with the final pooling layer replaced by a fully connected layer.
To determine the upper body region, we  conduct human pose estimation \cite{pose}
to identity keypoints of the body and the upper part is thereby located by these points.
The head, whole body and upper body models, which are used for COCO loss training, are finetuned on PIPA training set using the pretained inception model, following similar procedure of patch alignment stated previously for the face region.
The aligned patches of four regions are shown in Figure \ref{fig:pipeline}(c).

\subsection{Congenerous cosine loss for training}
\textcolor{black}{
The intuition behind designing a COCO loss is that we directly compare and optimize the cosine distance (similarity) between two features.
}
Let $\textbf{f}^{(i,r)}\in \mathbb{R}^{D \times 1}$ denote the feature vector of the $i$-th sample from region $r$, where $D$ is the feature dimension. For brevity, we drop the superscript $r$ since each region model undergoes the same COCO training.
We first define the \textit{cosine similarity} 
of two features 
from a mini-batch $\mathcal{B}$ 
as:
\begin{equation}
\mathcal{C}(\textbf{f}^{(i)}, \textbf{f}^{(j)}) =  
{ \textbf{f}^{(i)}  \cdot 
	\textbf{f}^{(j)}}  \big / {   \|\textbf{f}^{(i)} \|  
	\| \textbf{f}^{(j)}\| }.
\end{equation}
The cosine similarity measures how close two samples are in the feature space. A natural intuition to  a desirable loss is to increase the similarity of samples within a category and enlarge the centroid distance of samples across classes. Let $l_{i}, l_{j} \in \{1, \cdots, K\}$ be the labels of sample $i, j$, where $K$ is the total number of categories,
we have the 
following loss
to maximize:
\begin{gather}
\mathcal{L}^{naive} = \sum_{i,j \in \mathcal{B}} \frac{\delta(l_{i}, l_{j}) \mathcal{C}(\textbf{{f}}^{(i)}, \textbf{{f}}^{(j)}) }
{(1 - \delta(l_{i}, l_{j}) ) \mathcal{C}(\textbf{{f}}^{(i)}, \textbf{{f}}^{(j)}) + \epsilon}, \label{basic_loss}
\end{gather}
where $\delta(\cdot, \cdot)$ is an indicator function and $\epsilon$ is a trivial number for computation stability.  
Such a design is reasonable in theory and yet suffers from computational inefficiency. Since the complexity of the loss above is $\mathcal{O}(C_M^2)=\mathcal{O}(M^2)$, the loss increases quadratically as batch size $M$ goes bigger. Also the network  suffers from unstable parameter update and is hard to converge if we directly compute  loss from  two arbitrary samples from a mini-batch.

Inspired by the center loss \cite{center_loss}, we define the \textit{centroid} of class $k$ as the average of features over a mini-batch $\mathcal{B}$:
\begin{equation}
\textbf{c}_k = \frac{\sum_{i\in \mathcal{B}} \delta(l_{i}, k) \textbf{{f}}^{(i)} }{ \sum_{i\in \mathcal{B}} \delta(l_{i}, k) + \epsilon} \in \mathbb{R}^{D \times 1 }, \label{centroid_def}
\end{equation}
where $\epsilon$ is a trivial number for computation stability. Incorporating the spirit of Eqn. \ref{basic_loss} with class centroid, we have the following output of sample $i$ to maximize:
\begin{equation}
{p}_{l_i}^{(i)} = \frac{ \exp \mathcal{C}(\textbf{{f}}^{(i)}, \textbf{c}_{l_i}) }
{ \sum_{k \neq l_i} \exp \mathcal{C}(\textbf{{f}}^{(i)}, \textbf{c}_k) } \in \mathbb{R}. \label{coco_per_sample}
\end{equation}
The direct intuition behind Eqn. \ref{coco_per_sample} is to measure  the distance  of one sample against other samples by way of a class centroid, instead of a direct pairwise comparison as in Eqn. \ref{basic_loss}. 
The numerator ensures  sample $i$ is close enough to its own class $l_i$ and the denominator enforces a minimal distance against samples in other classes. The exponential operation is to transfer the cosine similarity 
to a normalized probability output, ranging from 0 to 1.

To this end, we propose the congenerous 
cosine (COCO) loss, which is to increase similarity within classes and enlarge variation across categories in a cooperative way:
\begin{gather}
	\mathcal{L}^{COCO} =  \sum_{i \in \mathcal{B}}	\mathcal{L}^{(i)} = - \sum_{k, i }    t_k^{(i)} \log p_{k}^{(i)} = -\sum_{i \in \mathcal{B}} \log p_{l_i}^{(i)}, 
	\label{coco_loss}
\end{gather}
where $k$ indexes along the class dimension in $\mathbb{R}^K$, $t_k^{(i)}$ is the binary mapping of sample $i$ based on its label $l_i$. In practice, COCO loss can implemented in a neat way via the softmax operation.
For Eqn. \ref{coco_per_sample}, if we constrain the feature and centroid to be normalized 
(\textit{i.e.}, $\hat{\textbf{f}} = {\textbf{f}} / { \| \textbf{f} \| }$, $\hat{\textbf{c}} = {\textbf{c}}/ { \| \textbf{c} \| }$) and loose the summation in the denominator to include $k=l_i$, the probability output of sample $i$ becomes:
\begin{equation}
 p_{k}^{(i)} = \frac{\exp (    \hat{ \textbf{c} }_{k}^{T} \cdot \hat{\textbf{f}}^{(i)} ) }{  \sum_m  \exp (\hat{ \textbf{c} }_{m}^{T} \cdot \hat{\textbf{f}}^{(i)}  )}
 =\texttt{softmax} \big(
 z_k^{(i)}
  \big),
 \label{softmax_form}
\end{equation}
where $z_k^{(i)}=\hat{ \textbf{c} }_{k}^{T} \cdot \hat{\textbf{f}}^{(i)}$ is the input to softmax. 
$\hat{ \textbf{c} }_{k}$ can be seen as  weights in the classification layer with  bias term being zero. \textcolor{black}{The advantage of COCO loss in Eqn. \ref{coco_loss} and \ref{softmax_form} from the naive version in Eqn. \ref{basic_loss} are two folds: it reduces the complexity of computation and could be achieved via the softmax with normalized inputs in terms of cosine distance.}

The derivative of loss $\mathcal{L}^{(i)}$ w.r.t. the input 
feature $\textbf{f}^{(i)}$,  
written in an element-wise form and dropping sample index $i$ for brevity, is as follows:
\begin{align}
	\frac{\partial \mathcal{L}}{ \partial f_j} & = \sum_d 
	\nabla_{d} \mathcal{L}
	\cdot 
	 \frac{ \partial \hat{f}_d} { \partial {f}_j}, \nonumber \\
& = \nabla_{j} \mathcal{L}
\cdot 
\frac{ \partial \hat{f}_j} { \partial {f}_j}    + \sum_{d \ne j}  
\nabla_{d} \mathcal{L}
\cdot 
\frac{ \partial \hat{f}_d} { \partial {f}_j}, \nonumber \\
	%
	%
& =  \frac {\nabla_{j} \mathcal{L}} {\| \textbf{f} \| } - \sum_{d} \nabla_{d} \mathcal{L} \cdot \frac{f_i  f_d}{  \| \textbf{f} \|^3  }, \nonumber\\
	%
	%
& =  \frac {
	\nabla_{j} \mathcal{L}        - (  \nabla \mathcal{L}^T \cdot \hat{\textbf{f}}    )  \hat{f}_j       
	} {\| \textbf{f} \| }, \\
\nabla_{x} \mathcal{L}   \triangleq 
  \frac{\partial \mathcal{L}}{ \partial \hat{f}_x}
& = \sum_k \frac{\partial \mathcal{L}}{ \partial z_k} \cdot \frac{ \partial z_k}{\partial \hat{f}_x},  \nonumber \\
& = \sum_k (p_k - t_k) \cdot \hat{c}_{kx},
\end{align}
where $\nabla \mathcal{L} \in \mathbb{R}^D$ is the top gradient w.r.t. the normalized feature. 
We  can derive the gradient w.r.t. centroid ${\partial \mathcal{L}}/{ \partial c_{kj}}$ in a similar manner.
Note that both the features and  cluster centroids 
are 
trained end-to-end. 
The features are initialized from the pretrain models and the initial value of $\textbf{c}_k$ is thereby obtained via Eqn. \ref{centroid_def}.

\begin{figure}[t]
		\centering
		\includegraphics[width=.45\textwidth]{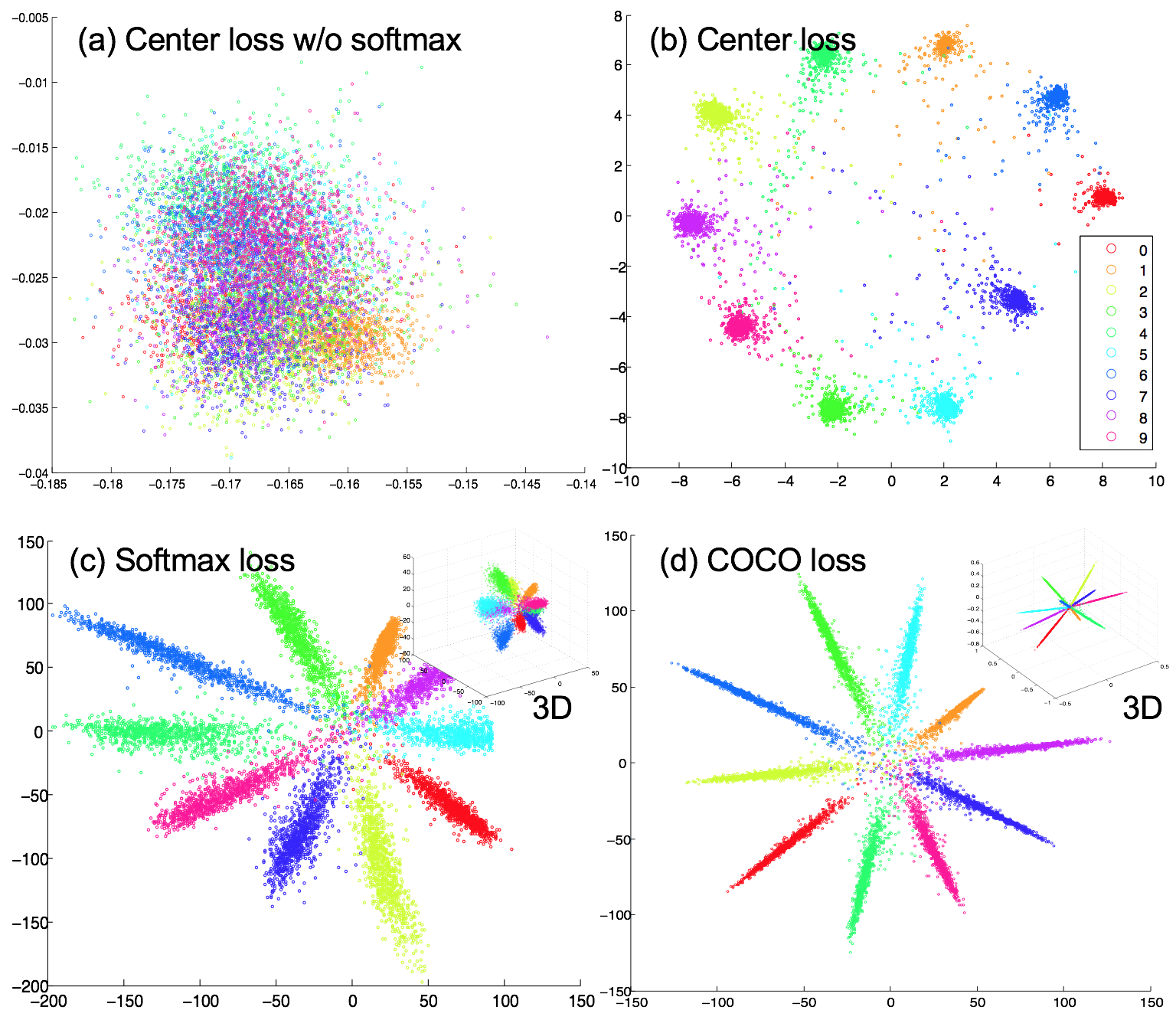}
	\cprotect \caption{Feature visualization using different losses, trained on MNIST \cite{mnist} with 10 classes. The softmax loss tries to chock up the feature space while COCO enlarges  inter-class distance as well as alleviates  inner-class variation. 3D cases in insets of (c) and (d) are more evident.
	}
	\label{fig:loss_motivation}
\end{figure}

\begin{figure*}
	\centering
	\includegraphics[width=.95\textwidth]{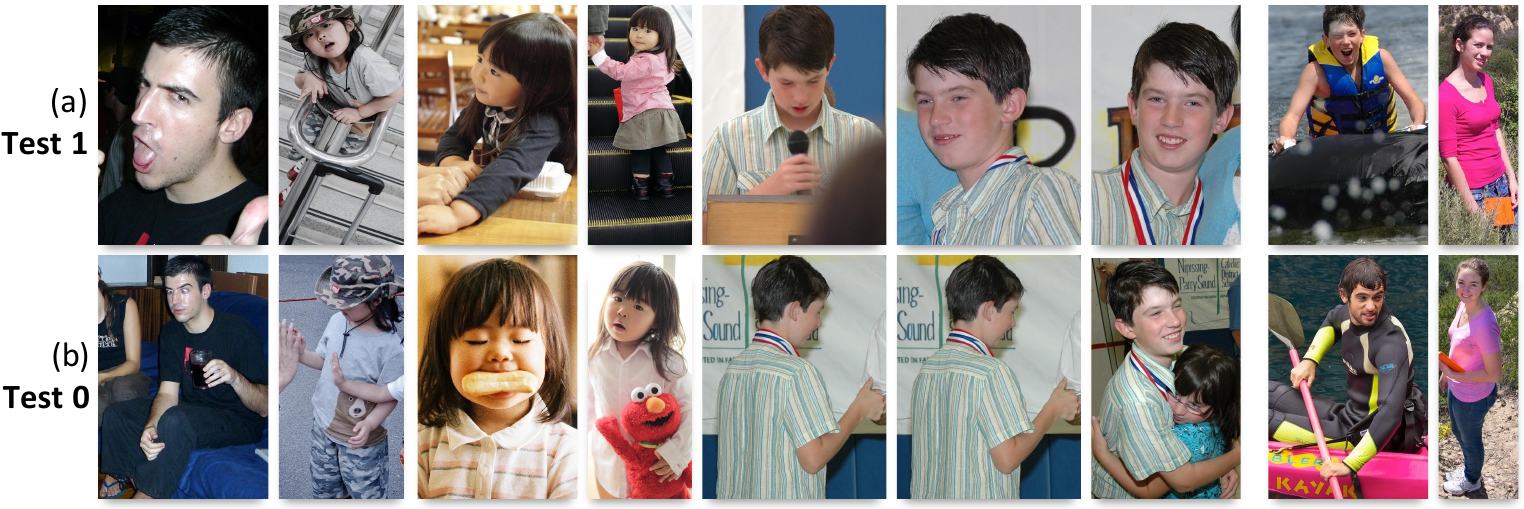}
	\cprotect \caption{Visualization of our method on the PIPA \texttt{test\_1} set. 
		Given the 
		image as input to be predicted in (a), its nearest neighbour in the feature space from  \texttt{text\_0} is shown in (b). The identity of the input is determined by the label of its neighbour.
		Our model can handle complex scenes with non-frontal faces and body occlusion. 
		The last two columns show failure cases.
		}
	\label{fig:visualize}
\end{figure*}

\subsection{Relationship of COCO with  counterparts}

COCO loss is formulated as a metric learning approach in the feature space, using cluster centroid in the cosine distance as metric to both enlarge inter-class variation as well as narrow down inner-class distinction. It can be achieved via a softmax operation under several constraints.
Figure \ref{fig:loss_motivation} shows the visualization of feature clusters under different loss schemes. 
For softmax loss, it only enforces samples across categories to be far away while ignores the similarity within one class (\ref{fig:loss_motivation}(c)). In COCO loss, we replace the weights in the classification layer before softmax, with a clearly defined and learnable cluster centroid (\ref{fig:loss_motivation}(d)).
The center loss \cite{center_loss} is similar in some way to ours. However, it needs an external memory to store the center of classes and thus the computation is twice as ours (\ref{fig:loss_motivation}(a-b)).
\textcolor{black}{
	\cite{quite_similar_loss_to_coco} proposed a generalized large-margin softmax loss, which also learns discriminative features by imposing intra-class compactness
and inter-class separability.}
The coupled cluster loss \cite{loss} is a further adaptation from the triplet loss \cite{triplet_loss_xiaolong} where in one mini-batch, the positives of one class will get as far as possible from the negatives of other classes. It optimizes the inter-class distance in some sense but fails to differentiate among the negatives. 

\begin{table*}[t]
	\begin{center}
		\caption{Ablation study on body regions and feature alignment using softmax as loss. We report the classification accuracy 
			(\%) 
			where \texttt{n-a} denotes the non-alignment case and \texttt{ali} indicates the alignment case.
			Note that for the face region, we only evaluate instances with faces.}
		\label{tab:region}
\vspace{-.2cm}
		\footnotesize{
			\begin{tabular}{c | c c | c c c | c c c | c c }
				\toprule
				\multirow{2}{*}{
				Test split } & \multicolumn{2}{c|}{Face} & \multicolumn{3}{c|}{Head} 
				& \multicolumn{3}{c|}{Upper body} & \multicolumn{2}{c}{Whole body}   \\
				& \texttt{n-a} & \texttt{ali} & \texttt{n-a} & \texttt{ali}& \cite{person_recog_rnn} & \texttt{n-a} & \texttt{ali}& \cite{person_recog_rnn} & \texttt{n-a} & \texttt{ali} \\
				\midrule 
				\texttt{ original} &  95.47 & 97.45 & 74.23 & 82.69  & 81.75 & 76.67 & 80.75  & 79.92 & 75.04 & 79.06 \\
				\texttt{ album}    & 94.66 & 96.57 & 65.47 & 73.77 & 74.21& 66.23 & 69.58 & 70.78 & 64.21 & 67.27  \\
				\texttt{ time}  & 91.03 & 93.36 & 55.88  & 64.31 & 63.73& 55.24  &  57.40  & 58.80& 55.53 & 54.62   \\
				\texttt{ day} & 90.36 & 91.32 & 35.27 & 44.24 & 42.75 & 26.49  & 32.09 &34.61& 32.85 & 29.59   \\
				\bottomrule
			\end{tabular}
		}
	\end{center}
\end{table*}

\subsection{Inference}\label{sec:inference}
At testing stage, we measure the similarity of features between two test splits to recognize the identity of each instance in \texttt{test\_1} based on the labels in \texttt{test\_0}.
The similarity between two patches $i$ and $j$ in \texttt{test\_1} and \texttt{test\_0} is denoted by
$s^{(r)}_{ij} = \mathcal{C}(\textbf{f}^{(i, r)}, \textbf{f}^{(j, r)})$, where $r$ indicates a specific region
model. A key problem is how to merge the similarity scores from different regions. We first normalize the preliminary result $s^{(r)}$ in order to have scores across different regions comparable:
\begin{gather}
\hat{s}^{(r)}_{ij}  = \bigg( 1 + \exp \big[- (  \beta_0 + \beta_1 s^{(r)}_{ij}  )  \big]  \bigg)^{-1},
\end{gather}
where $\beta_0, \beta_1$ are parameters of the logistic regression.
The final score $S_{ij}$ is a weighted mean of the normalized scores $\hat{s}^{(r)}_{ij} $ of each region:
$S_{ij} = \sum_{r=1}^{R} \gamma^r \cdot \hat{s}^{(r)}_{ij}$, where
$R$ is the total number of regions and $\gamma^r $ being the weight of each region's score.
The identity of patch $i$  in \texttt{test\_1} is decided by the label corresponding to the maximum score in the reference set:
$l_i = \arg \max_{j*} S_{ij}$.
Such a scheme guarantees that when new training data are added into \texttt{test\_0}, there is no need to train a second model or SVM on the reference set, which is quite distinct from previous work. 
\textcolor{black}{The test parameters of $\beta$ and $\gamma$ are determined by a validation set on PIPA.}

\begin{table*}[t]
	\begin{center}
		\caption{Investigation on the combination of merging  similarity score from different body regions during inference. The top two results in each  split are marked in  {}{\textbf{bold}} and {}{\textit{\underline{italic}}}. COCO loss is applied  in all cases except the last one with softmax loss.}  \label{tab:score_merge}
		\vspace{-.2cm}
		\footnotesize{
			\begin{tabular}{c | c c c c | c c c c}
				\toprule
				Method & Face  & Head &  Upper body& Whole body & \texttt{original} & \texttt{album} & \texttt{time} &\texttt{day}
				\\
				\midrule
				-& \checkmark & \checkmark & && 84.17 & 80.78 & 74.00 & {}{{53.75}}\\
				-& \checkmark & & \checkmark && 89.24 & 81.46 & \underline{\textit{76.84}} & {}{\textit{\underline{61.48}}}\\
				-& \checkmark && & \checkmark & 88.40 & 82.15 & 70.90 & 57.87\\
				-& & \checkmark &\checkmark  && 88.76 & 79.15 & 68.64 & 42.91\\
				\cite{person_recog_rnn} & & \checkmark &\checkmark  && 84.93 & 78.25 & 66.43 & 43.73\\
				-& & \checkmark & &\checkmark & 87.43 & 77.54 & 67.40 & 42.30\\
				-& & & \checkmark & \checkmark & 81.93 & 73.84 & 62.46 & 34.77 \\
				\midrule
				-& \checkmark & \checkmark & \checkmark&& 87.86 & 80.85 & {}{{{71.65}}} & 59.03 \\
				-& \checkmark & \checkmark & &\checkmark & {}{{{88.13}}} &{}{\textit{\underline{82.87}}} &73.01 &55.52\\
				-& & \checkmark & 	\checkmark &	\checkmark & 89.71 & 78.29 & 66.60 & 52.21 \\
				-& \checkmark & &\checkmark  &\checkmark & \underline{\textit{91.43 }}& 80.67 & 70.46 & 55.56  \\
				\midrule
				-& \checkmark & \checkmark & \checkmark &\checkmark & {}{\textbf{92.78}} & {}{\textbf{83.53}} & {}{\textbf{77.68}} & \textbf{61.73} \\
				Softmax& \checkmark & \checkmark & \checkmark &\checkmark & {}{{88.73}} & {}{{80.26}} & {}{{71.56}} & 50.36 \\
				\bottomrule
			\end{tabular}
		}
	\end{center}
\end{table*}

\begin{table}
	\begin{center}
		\cprotect\caption{Recognition accuracy (\%) comparison of our approach with state-of-the-arts on PIPA.
			} \label{tab:compare}
		\vspace{-.2cm}
		\footnotesize{
			\begin{tabular}{c c c c c}
				\toprule
				Methods & \texttt{ original} & \texttt{ album} & \texttt{ time}  & \texttt{ day}
				\\
				\midrule
				PIPER
				& 83.05 & - & - & -\\
				RNN
				& 84.93 & 78.25 & 66.43 & 43.73 \\
				Naeil
				& 86.78 & 78.72 & 69.29 & 46.61 \\
				\midrule
				Ours & \textbf{92.78} & \textbf{83.53 }& \textbf{77.68 }& \textbf{61.73 }\\
				\bottomrule
			\end{tabular}
		}
	\end{center}
	
\end{table}

\section{Experiments}

\subsection{Setup}
\textbf{Dataset and evaluation metric.
}\textit{The People   In Photo Albums (PIPA)} dataset \cite{piper} is adopted for evaluation. The PIPA dataset is divided into train, validation, test and leftover sets, where the head of each instance is annotated in all sets.
The test set
is split into two subsets, namely \texttt{test\_0} and \texttt{test\_1} with roughly the same number of instances. Such a division is called the \texttt{original}  split.
As did in \cite{oh_iccv,person_recog_rnn,piper,zlin}, the training set
is only used for learning feature representations;
the recognition system is trained on \texttt{test\_0} and evaluated on \texttt{test\_1}. In this work, as mentioned previously,
we take full advantage of the training set and remove the second training on \texttt{test\_0}.
Moreover, \cite{oh_iccv} introduced three more challenging splits, 
including \texttt{album}, \texttt{time} and \texttt{day}. Each case emphasizes different temporal distance (different albums, events, days, etc.)
between the two subsets.
The evaluation metric is the averaged classification accuracy over all instances on \texttt{test\_1}.

\textbf{Implementation details.} For the face model, the initial learning rate is set to 0.001 and decreased by 10\% after 20 epochs. For other three models, 
the initial learning rate is set to 0.005 and decreased by 20\% after 10 epochs.
The weight decay and momentum are 0.005 and 0.9 across models. We use stochastic gradient descent with the Adam optimizer \cite{adam_opt}.
\textcolor{black}{Note that during training, each patch on region $r$ of an identity is cropped from the input, where the image for a specific person 
is resized to the extent that 
the longer dimension of the head is fixed at 224 across patches, ensuring the scale of different body regions for each instance is the same. 
Moreover, for the whole body model we 
simply use the similarity transformation (scale, transform, rotation) instead affine operation for better performance. }

\subsection{Component analysis}

\textbf{Feature alignment in different regions.} Table \ref{tab:region} reports the performance of using feature alignment and different
body regions, where several remarks could be observed. First, the alignment case in each region performs better by a large margin than the non-alignment case, which verifies the motivation of patch alignment to alleviate inner-class variance stated in Section \ref{sec:face-detection-and-alignment}. Second, for the alignment case, the most representative features to identify a person reside in the region of face, followed by head, upper body and whole body at last.
Such a clue is not that obvious  for the non-alignment case. Third, we notice that for the whole body region,  accuracy in the non-alignment case is higher than that of the alignment case in  \texttt{time} and \texttt{day}. This is probably due to the improper definition of base  points on these two sets.

\begin{figure}
	\centering
	\includegraphics[width=0.46\textwidth]{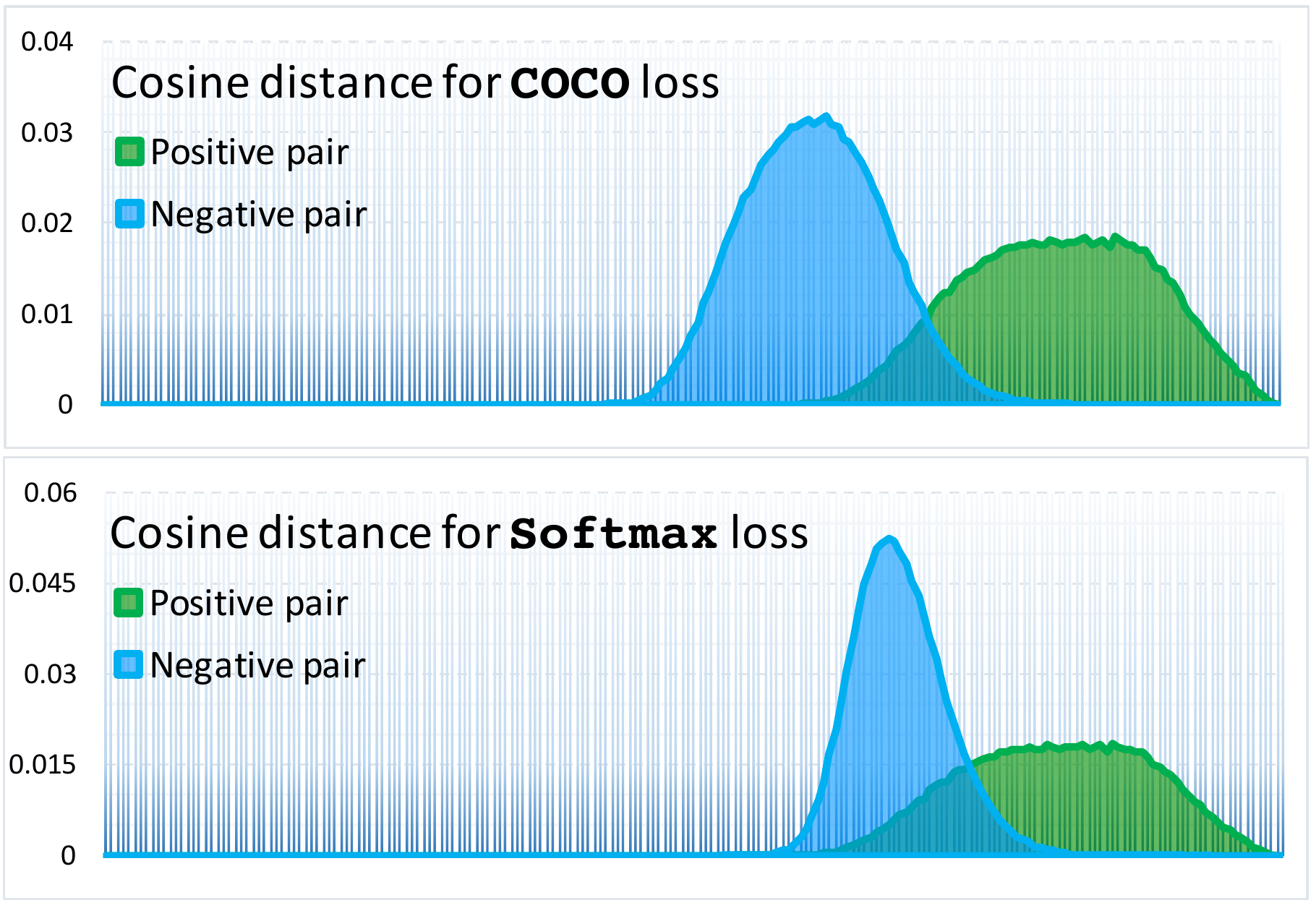}
	\vspace{-.1cm}
	\caption{Histogram of  the cosine distance for positive and negative pairs during inference using different losses.}
	\label{fig:roc}
\end{figure}

\textbf{Congenerous cosine loss.} Figure \ref{fig:roc} shows the histogram of the cosine distance among positive pairs (\textit{i.e.}, same identity in \texttt{test\_0} and \texttt{test\_1}) and negative pairs.
We can see that in the COCO trained model, the discrepancy between inter-class (blue) and inner-class (green) samples in the test set is well magnified; whereas in the softmax trained case, such a distinction is not obvious. This verifies the effectiveness of our COCO loss.

\textbf{Similarity integration from regions.} Table \ref{tab:score_merge} depicts the ablation study on the combination of merging the similarity score  from different body regions during inference. 
Generally speaking, taking all  regions into consideration could result in the best accuracy of 92.78 on the \texttt{original}  set. 
It is observed that the performance is still fairly good on  \texttt{day}  and  \texttt{time}  if the two scores of face and upper body alone are merged. 
%


\textcolor{black}{\cite{person_recog_rnn} also employs a  multi-region processing step and we include the comparison in Table \ref{tab:region} and \ref{tab:score_merge}.
On some splits our model is superior (\textit{e.g.}, head region, 82 vs 81 on \texttt{original}, 44 vs 42 on \texttt{day}); whereas on other splits ours is inferior (\textit{e.g.}, upper body region, 69 vs 70 on \texttt{album}, 57 vs 58 on \texttt{time}). This is probably due to distribution imbalance among 
splits: the upper body region differs greatly in appearance with some instances absent of this region, making COCO  hard to learn  features. However, under the score integration scheme, the final merged prediction can complement features learned among different regions and achieves better performance against \cite{person_recog_rnn}.}

\subsection{Comparison to state-of-the-arts}

We can see from Table \ref{tab:compare} that our recognition system outperforms against previous state-of-the-arts, PIPER \cite{piper}, RNN \cite{person_recog_rnn}, Naeil \cite{oh_iccv}, in all four test splits.
Figure \ref{fig:visualize} visualizes a few examples of the predicted instances by our model, where complex scenes with non-frontal faces and body occlusion can be handled properly in most scenarios. 
%
Failure cases are 
probably due to the almost-the-same appearance configuration in these scenarios.

\section{Conclusion}

In this work, we propose a person recognition method to identify the same person, where four models for different body regions are trained.
 Region patches are further aligned via affine transformation to make the model less prone to overfitting. Moreover, the training procedure
 employs a COCO loss to reduce the inner-class variance as well as enlarge inter-class varation.
Our pipeline requires only one-time training of the model; we utilize the similarity between \texttt{text\_0} and \texttt{text\_1} to determine the person's identity during inference.  
Experiments show that the proposed method outperforms against other state-of-the-arts on the PIPA dataset.

\section*{Acknowledgements}
We would like to thank reviewers for helpful comments; Hasnat, Weiyang and Feng for method discussion and paper clarification in the initial release. H. Li is financially supported by the Hong Kong Ph.D Fellowship Scheme.
\bibliographystyle{named}
\bibliography{icme2017template,deep_learning}


\end{document}